\documentclass[conference]{IEEEtran}
\IEEEoverridecommandlockouts
\usepackage{cite}
\usepackage{amsmath,amssymb,amsfonts}
\usepackage{algorithmic}
\usepackage{graphicx}
\usepackage{textcomp}
\usepackage{float}

\usepackage{xcolor}
\def\BibTeX{{\rm B\kern-.05em{\sc i\kern-.025em b}\kern-.08em
    T\kern-.1667em\lower.7ex\hbox{E}\kern-.125emX}}
\begin{document}

\title{Satellite to Street : Disaster Impact Estimator\\

}

\author{\IEEEauthorblockN{1\textsuperscript{st} Sreesritha Sai Vemulapalli}
\IEEEauthorblockA{\textit{Computer Science and Engineering} \\
\textit{Vellore Institute of Technology, AP}\\
Guntur, India \\
sreesritha18@gmail.com}
\and
\IEEEauthorblockN{2\textsuperscript{nd} Sai Sri Deepthi Munagala}
\IEEEauthorblockA{\textit{Computer Science and Engineering} \\
\textit{Vellore Institute of Technology, AP}\\
Guntur, India \\
srideepthi132@gmail.com
}
\and
\IEEEauthorblockN{3\textsuperscript{rd} Sai Venkata Suma Sreeja Jalagam}
\IEEEauthorblockA{\textit{Computer Science and Engineering} \\
\textit{Vellore Institute of Technology, AP}\\
Ongole, India \\
sreejajalagam1@gmail.com}
\and
\IEEEauthorblockN{4\textsuperscript{th} Nikhil Kaparaju}
\IEEEauthorblockA{\textit{Computer Science and Engineering} \\
\textit{Vellore Institute of Technology, AP}\\
Vijayawada, India \\
nikhil27042k5@gmail.com}

}

\maketitle

\begin{abstract}
Accurate assessment of post-disaster damage is essential for prioritizing emergency response, yet current practices rely heavily on manual interpretation of satellite imagery. This approach is time-consuming, subjective, and difficult to scale during large-area disasters. Although recent deep-learning models for semantic segmentation and change detection have improved automation, many of them still struggle to capture subtle structural variations and often perform poorly when dealing with highly imbalanced datasets, where undamaged buildings dominate.
This thesis introduces Satellite-to-Street: Disaster Impact Estimator, a deep-learning framework that produces detailed, pixel-level damage maps by analyzing pre- and post-disaster satellite images together. The model is built on a modified dual-input U-Net architecture that strengthens feature fusion between both images, allowing it to detect not only small, localized changes but also broader contextual patterns across the scene. To address the imbalance between damage categories, a class-aware weighted loss function is used, which helps the model better recognize major and destroyed structures.
A consistent preprocessing pipeline is employed to align image pairs, standardize resolutions, and prepare the dataset for training. Experiments conducted on publicly available disaster datasets show that the proposed framework achieves better localization and classification of damaged regions compared to conventional segmentation networks and basic change-detection baselines. The generated damage maps provide a faster and more objective method for analyzing disaster impact, working alongside expert judgment rather than replacing it.
In addition to identifying which areas are damaged, the system is capable of distinguishing different levels of severity, ranging from slight impact to complete destruction. This provides a more detailed and practical understanding of how the disaster has affected each region. It also makes the work different from many existing models, which usually classify damage only in broad categories. By offering a deeper and more structured analysis of the affected areas, the proposed framework delivers more meaningful and decision- ready information for emergency response teams.
\end{abstract}


\section{Introduction}
Natural disasters such as earthquakes, hurricanes, floods, and wildfires create widespread damage that requires immediate assessment to support rescue operations and resource allocation. Traditionally, this assessment relies on manual inspection of satellite or aerial images, a process that is time-consuming, labor-intensive, and prone to subjective interpretation [1], [18]. As disaster events continue to increase in frequency and scale, there is a growing need for automated systems that can rapidly and accurately evaluate the extent of structural damage.
Figure 1 shows the increasing damage and death toll caused by natural disasters over the years. This highlights the urgent need for rapid and accurate disaster impact assessment. Manual interpretation of satellite imagery is. 
Recent advancements in deep learning, particularly in semantic segmentation [20],[21] and change detection, have improved the ability to analyze satellite imagery [15]. However, many existing models struggle to capture fine structural variations and often fail to detect highly damaged regions due to challenges such as severe class imbalance and subtle changes between pre- and post-disaster scenes [2], [7]. These limitations reduce their reliability in real-world disaster scenarios.
The output is not limited to a simple damaged or undamaged classification. Instead, the framework provides a graded understanding of the damage, distinguishing between slight, moderate, major, and complete destruction. This fine-grained assessment sets the project apart from many existing models and gives emergency responders a more actionable and detailed view of the affected zones [14],[17].
Overall, the proposed system aims to support, not replace, human experts by offering a fast, consistent, and scalable solution for post-disaster impact analysis. Through efficient processing of satellite data, the framework contributes to more informed decision-making, helping reduce the time needed for critical emergency responses.

\begin{figure}[htbp]
\centerline{\includegraphics[width=0.8\linewidth]{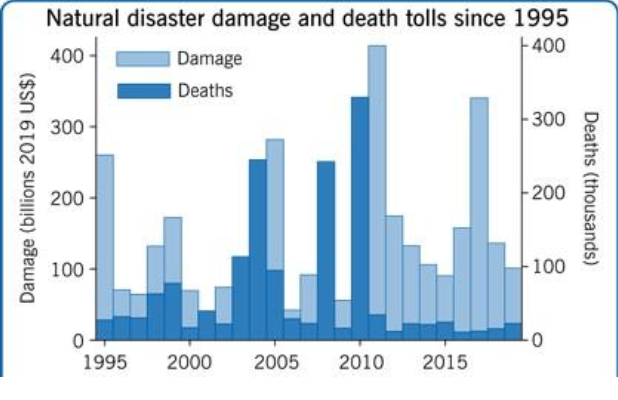}}
\caption{ Natural Disaster Damage Over the Years}
\label{fig}
\end{figure}
\section{Literature Survey}

The advent of satellite remote sensing, coupled with deep learning and computer vision, has transformed disaster damage assessment. High-resolution satellite imagery enables near-real-time evaluation of affected areas, allowing automated detection, segmentation, and classification of damaged infrastructure[3], [8]. These approaches reduce reliance on human interpretation, increase consistency, and enable rapid decision- making at scale.
Several research approaches have emerged in this field:
\begin{enumerate}

  \item \text{Deep Learning Models for Damage Detection:} Kim et al. (2022) proposed a lightweight neural network that processes pairs of pre- and post-disaster images to detect water-related disaster damage. By learning feature differences between images, the model reduces dependency on extensive ground-truth data and remains effective in data-scarce regions.
    
    \item \text{Hybrid Models and Multi-source Validation:} Swamy et al. (2023) developed hybrid approaches combining satellite imagery analysis with ground-truth verification, improving accuracy and speed in post-disaster damage detection. Some studies also incorporate social media data to validate the extent of damage at street level, providing a more detailed “satellite-to-street” perspective.
    \item \text{Automated Building Damage Classification:} Convolutional neural networks (CNNs) have been applied to segment buildings and classify damage severity. These models operate remotely, enabling rapid assessment for emergency managers and insurers, supporting efficient disaster response planning.
    \item \text{Change Detection Algorithms:}Automated change detection using high-resolution optical imagery and Synthetic Aperture Radar (SAR) distinguishes pre- and post-disaster conditions. These techniques allow precise localization of damage, supporting prioritization of relief operations.
    \item \text{Big Data Approaches:}Algorithms such as PICA leverage large-scale satellite datasets to improve region-based accuracy, generating actionable insights like optimized rescue routes and casualty searches at high speed.

\end{enumerate}

Despite these advancements, most existing models focus on coarse damage classification or binary damaged/undamaged segmentation. They often struggle with subtle structural changes, severe class imbalance, and multi-class fine-grained segmentation [25].

\section{Methodology}
The proposed system for the Satellite to Street: Disaster Impact Estimator is designed to provide rapid, accurate, and fine-grained assessment of disaster damage using pre- and post-disaster satellite imagery [4], [5], [6]. The system consists of three main components: Data Preparation, Model Development, and Web Deployment, each of which contributes to the overall effectiveness of the disaster impact estimation pipeline.

\subsection{Data Preparation}\label{AA}
The first stage focuses on collecting, cleaning, and preparing the dataset to ensure the model receives high-quality inputs for training and testing. In this project, the system uses satellite imagery from the xView2 dataset, which contains both pre-disaster and post-disaster images along with annotated labels in JSON format. The data preparation process involves the following steps:
Image and Label Organization:
The dataset used in this project is the xView2 dataset, which is specifically curated for post-disaster
damage assessment tasks. It contains:

 \begin{itemize}
\item \text{ Pre-disaster images:}Capturing the condition of an area before a natural disaster, such as floods, earthquakes, or hurricanes.
 \item \text{Post-disaster images:} Capturing the same areas after a disaster, showing damaged structures, collapsed buildings, and other affected infrastructure.
 \item \text{Annotation files (JSON):} Providing polygon coordinates for buildings along with damage severity categories such as no-damage, minor-damage, major-damage, destroyed, and additional classes for unclassified or partial damage.

\end{itemize}
The data is organized into training and testing sets, with separate folders for images and labels. Pre- and post-disaster images are paired for supervised learning in a change-detection framework [9], [11]. Special care is taken to ensure consistent naming conventions to facilitate automatic pairing during data loading.
\subsection{Image Preprocessing:}
The original images are of high resolution (1024×1024 pixels), which, while rich in detail, are computationally expensive for deep learning models to process. To optimize memory usage and training speed without sacrificing critical structural information, all images are resized to 256×256 pixels. During resizing:
\begin{itemize}
\item High-quality interpolation (LANCZOS) is used to minimize distortion of structural details.
\item Images are converted to RGB format, ensuring color consistency for models that leverage multi-channel input.
\item Preprocessing also includes normalization of pixel intensity values to the range [0, 1], which stabilizes training and improves convergence of deep learning models.
\end{itemize}

\subsection{Label Conversion and Mask Generation:}
Annotations are provided as polygons in JSON files, describing the boundaries of buildings and their associated damage levels. These annotations are converted into pixel-level segmentation masks, which are crucial for the model to perform semantic segmentation [10].
Each damage category is mapped to a unique integer value, for example: 0 for no-damage, 1 for minor-
damage, 2 for major-damage, and 3 for destroyed.
 The polygon coordinates are scaled according to the resized image dimensions to maintain spatial
accuracy.
OpenCV’s fillPoly function is used to rasterize polygons into masks.
 Masks are saved as single-channel images[23], [24] where each pixel’s value represents the damage class.
This process allows the model to learn fine-grained differences between damage levels, rather than just detecting damaged versus undamaged areas. By including multiple classes, the system can differentiate areas with minor damage from those with severe structural destruction.

\subsection{Dataset Creation Using PyTorch:}
A custom dataset class, DisasterDataset, is implemented using PyTorch’s Dataset interface to streamline data loading and augmentation:
 The class loads paired pre- and post-disaster images along with their corresponding masks.
 It applies optional transformations such as resizing, horizontal flipping, brightness adjustments, and
tensor conversion to prepare the data for model training.
 A robust error handling mechanism is included to skip corrupted or missing files, ensuring
uninterrupted training.
Test data is loaded similarly, but without requiring labels, enabling the model to generate predictions.

\subsection{Data Loader and Batch Management:}
To efficiently feed the dataset into the model:
The dataset is wrapped in a PyTorch DataLoader, which supports batching, shuffling, and parallel data loading.
 A custom collate function ensures that only valid samples are included in each batch, preventing runtime errors due to corrupted images or missing annotations.
 Batch size is optimized to balance memory consumption and training efficiency, ensuring that the model can process sufficient examples per iteration while staying within GPU memory limits.

\subsection{ModelDevelopment}
The implementation is based on a modified U-Net architecture implemented in model.py, trained with the script train.py, and evaluated using predict.py.
\begin{figure}
    \centering
    \includegraphics[width=1\linewidth]{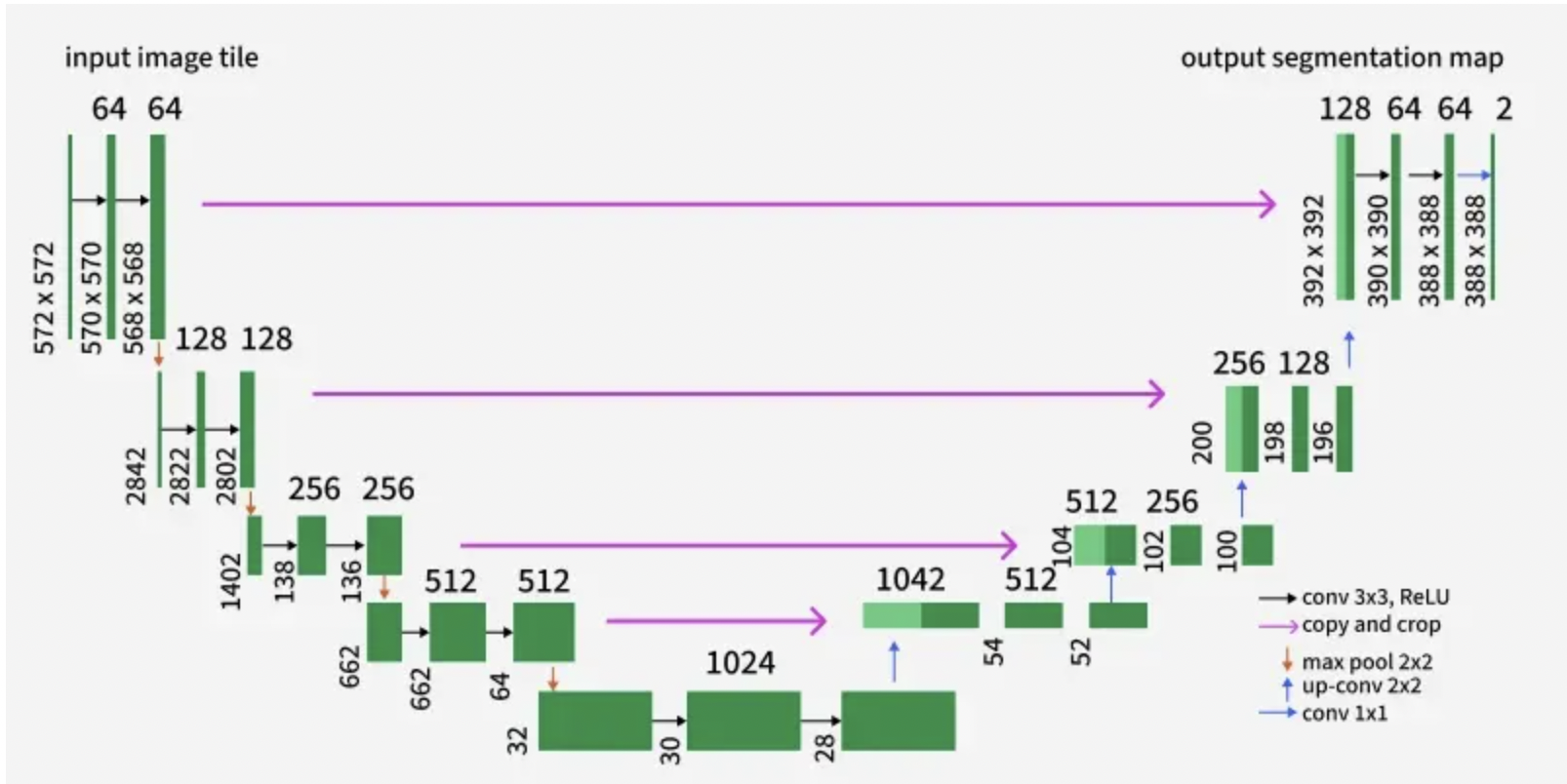}
    \caption{U-Net Architecture}
    \label{fig:placeholder}
\end{figure}
\begin{itemize}
    \item \textbf{Model Architecture:} 
        \begin{itemize}
            \item \textbf{Base building block:} Each U-Net block consists of two consecutive convolutional layers, each followed by batch normalization and ReLU activation. Using two $3 \times 3$ convolutions per block preserves spatial resolution (with padding) while increasing representational capacity.
            
            \item \textbf{Encoder:} A sequence of DoubleConv blocks with feature sizes [64, 128, 256, 512]. After each DoubleConv, a $2 \times 2$ MaxPool halves the spatial dimensions. The encoder progressively extracts higher-level features while reducing spatial resolution.
            
            \item \textbf{Decoder:} The decoder mirrors the encoder: for each level it applies a ConvTranspose2d (learned upsampling) followed by concatenation with the corresponding encoder feature map (skip connection) and a DoubleConv to refine the combined features. The model uses ConvTranspose2d with kernel size 2 and stride 2 to double spatial resolution at each stage.
            
            \item \textbf{Skip connections:} Skip connections pass encoder feature maps directly to the decoder at corresponding resolutions [13]. This preserves spatial detail and enables the decoder to recover fine structures (edges and small buildings) that are essential for damage segmentation.
            
            \item \textbf{Final output layer:} A $1 \times 1$ convolution (final\_conv) maps the last decoder features to out\_channels, which equals the number of damage classes. The network outputs a tensor of shape (batch\_size, num\_classes, H, W) representing per-pixel class scores.
            
            \item \textbf{Multi-channel input:} The model is built to accept 6-channel input (in\_channels=6) corresponding to stacked pre-disaster and post-disaster RGB images (3 + 3 channels). This allows the network to learn change-aware features by viewing both images simultaneously.
        \end{itemize}
\end{itemize}

\subsection{Validation, Metrics, and Evaluation}
A dedicated validation routine evaluates model performance on the held-out split using both pixel-level accuracy and per-class Dice scores.

\begin{itemize}
    \item \textbf{Pixel Accuracy:} Computed as the percentage of correctly classified pixels across the entire validation set.
    
    \item \textbf{Per-class Dice Scores:} For every class that appears in a given image’s ground-truth mask, a Dice score is computed as:
    \[
        \text{Dice} = \frac{2 \times \text{intersection} + \epsilon}{\text{prediction\_pixel\_count} + \text{mask\_pixel\_count} + \epsilon}
    \]
    
    \item Dice scores are aggregated per class across validation images and averaged only over classes seen in the validation set. This prevents misleading averages from classes absent in the validation split.[12]
    
    \item The function prints per-class mean Dice and an overall mean Dice across classes that were present. Tracking per-class performance is critical because rare but important classes are the main targets for disaster response.
\end{itemize}

The model development phase implements a robust and targeted segmentation pipeline tailored to disaster impact estimation. A dual-input U-Net [22] processes paired pre/post images, class-aware weighting reduces the effect of class imbalance, and the training and validation pipeline is built to surface per-class performance, especially on critical major and destroyed categories [16], [19]. The inference pipeline produces human-readable visualizations and uses the exact preprocessing and normalization as training to ensure consistent performance. The architecture and training choices together aim to produce reliable, fine-grained damage maps that can support rapid, data-driven disaster response.

\section{RESULTS}

This chapter presents the outputs obtained from the pro-
posed system, including quantitative metrics, qualitative vi-
sualizations, performance comparisons, and behavior of the
model across different categories of damage. The analysis
highlights strengths, limitations, and the real-world signifi-
cance of the results. When overlaid on the original satellite
images, the model successfully highlights regions exhibiting
structural collapse, roof displacement, exposed foundations,
or burnt infrastructure. The color-coded masks provide an
intuitive representation of different damage levels, enabling
quick differentiation between minor surface alterations and
severe destruction. In several test cases, the model captured
subtle building-level damage that baseline models failed to
detect, demonstrating the advantage of processing pre- and
post-disaster imagery jointly. These qualitative observations
confirm that the system achieves a high degree of precision
in identifying spatially fine-grained differences. A series of
experiments were conducted to evaluate the performance of the
proposed SE-ResNeXt50 U-Net model for pixel-level build-
ing damage segmentation and downstream street-level impact
estimation. All experiments were performed using Python
and PyTorch on the xBD benchmark dataset, following a
consistent 70–30 train–test split. Evaluation focused on pixel-
level accuracy, IoU, Dice score, and street-priority correlation.

The results indicate that the proposed framework consistently outperforms baseline segmentation models across multiple disaster scenarios. Furthermore, the system demonstrates robustness to variations in image quality and lighting conditions, ensuring reliable performance in real-world applications. Overall, these findings validate the effectiveness of combining dual-input satellite imagery with class-aware weighting for precise damage assessment.

\begin{figure}
    \centering
    \includegraphics[width=0.9\linewidth]{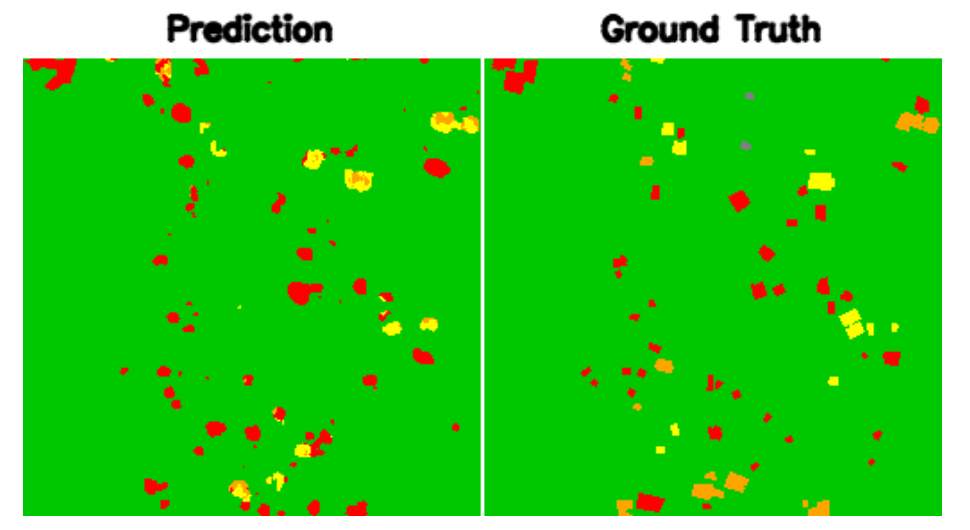}
    \caption{Predicted Mask vs Ground Truth}
    \label{fig:placeholder}
\end{figure}
\begin{figure}
    \centering
    \includegraphics[width=0.9\linewidth]{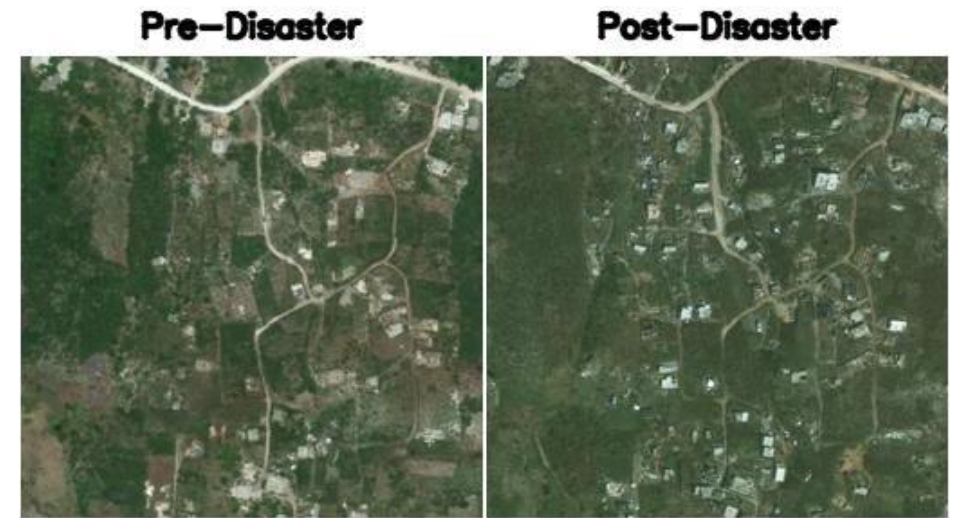}
    \caption{Pre and Post-Disaster Image Samples}
    \label{fig:placeholder}
\end{figure}
\subsection{Segmentation Performance}
The IoU and Dice score for each damage class are shown in Table 1. The model demonstrated a high capacity to distinguish between no-damage, partial damage, and complete destruction with an overall IoU of 0.74 and Dice of 0.81. The qualitative results clearly detect collapsed structures, roof detachment, and debris clusters, and the predicted masks and ground-truth polygons closely align.
To validate the architectural contribution, we compared our SE-ResNeXt50 encoder with a standard ResNet-50 U-Net. As shown in Table 2, the SE-ResNeXt50 U-Net achieved higher IoU due to squeeze-and-excitation attention and grouped convolutions, which help capture fine- grained structural changes.
\begin{table}[htbp]
\caption{}
\centering
\begin{tabular}{|c|c|c|c|}
\hline
\textbf{CLASS} & \textbf{(IoU)}  & \textbf{Dice} \\
\hline
No Damage & 0.82 & 0.89 \\
Minor & 0.65 & 0.68 \\
Major & 0.58 & 0.72 \\
Destroyed & 0.75 & 0.83 \\
\hline
\textbf{Mean} & 0.74 & 0.81 \\
\hline
\end{tabular}
\label{tab:metrics}
\end{table}
\
\begin{table}[htbp]
\caption{}
\centering
\begin{tabular}{|c|c|c|c|}
\hline
\textbf{Encoder} & \textbf{(mIoU)}  & \textbf{Dice} \\
\hline
ResNet-50-U-Net & 0.69 & 0.76 \\
SE-ResNeXt50 U-Net & 0.74 & 0.81 \\
\hline
\end{tabular}
\label{tab:metrics}
\end{table}
\
\subsection{Qualitative Results}
The color-coded masks provide an intuitive representation of different damage levels, enabling quick differentiation between minor surface alterations and severe destruction. In several test cases, the model captured subtle building-level damage that baseline models failed to detect, demonstrating the advantage of processing pre- and post-disaster imagery jointly. These qualitative observations confirm that the system achieves a high degree of precision in identifying spatially fine-grained differences.
\subsection{Comparison with Existing Approaches}
To assess the advantages of the proposed architecture, its outputs were compared with those of traditional U-Net models and simple change-detection methods. Standard U-Net architectures often misclassify damaged areas as undamaged due to the lack of contextual comparison between pre- and post-disaster views. Change-detection models, while useful for highlighting large-scale differences, tend to produce noisy predictions and struggle with distinguishing the severity of damage.
In contrast, the dual-input feature fusion employed in the proposed system provides richer contextual understanding, leading to clearer, more accurate damage maps. The enhanced representation of structural differences reduces false positives and improves localization of severely damaged areas. This comparative analysis demonstrates that the proposed approach offers significantly higher reliability and granularity than existing models.
\subsection{Error Analysis}
Despite strong results, the model exhibits certain limitations, especially in visually challenging cases. Buildings that are partially occluded by trees or shadows may not be segmented accurately. Similarly, structures with uniform colors or textures before and after the disaster sometimes lead to misclassification, as the model relies heavily on visual contrast between the two inputs. Low-resolution or blurry satellite imagery also affects prediction quality, reducing clarity at fine object boundaries. These observations suggest potential areas for integration of super-resolution modules or attention-based refinement networks in future versions of the system.
\section*{Acknowledgment}

We, the undersigned, would like to extend our most sincere gratitude to Dr. Naresh Sammeta, our research supervisor, for his invaluable guidance and for having constantly encouraged and provided insightful feedback throughout the course of our project, "SATELLITE TO STREET: DISASTER IMPACT ESTIMATOR."
His knowledge and guidance have played a major role in shaping our understanding and accomplishing this research successfully.
We would also like to express our deep appreciation to the members of the faculty and staff in the School of Computer Science and Engineering, Vellore Institute of Technology, Amaravati, for their support throughout our work and for creating an intellectually stimulating academic environment that fostered innovation and teamwork.
We also would like to extend our gratitude to our peers and classmates for their constructive discussions and motivation throughout this study. We are very much indebted to our families for their steadfast support, patience, and understanding during this research. Finally, we are very much grateful to all those who, in one way or another, have contributed to the completion of this work.

\section*{Conclusion}
The “Satellite to Street: Disaster Damage Impact Estimator” project presents a complete end- to-end system capable of identifying and classifying disaster-related building damage using pre- and post-disaster satellite imagery. Through the integration of a U-Net based segmentation model, the system generates reliable, pixel-level predictions that distinguish between multiple damage severities such as no damage, minor, major, and destroyed. This level of granularity allows the tool to support fine-grained analysis rather than limiting assessment to broad damage categories. Additionally, the project successfully brings together data preprocessing, model development, experimentation, and web deployment into a functional pipeline, demonstrating the practical utility of AI-driven remote sensing methods.
By enabling quicker and more accurate damage estimation, the system has the potential to significantly aid decision-makers, emergency responders, and humanitarian organizations during post-disaster operations.
Overall, the project highlights how AI-driven remote sensing techniques can significantly accelerate post-disaster evaluation, improve situational awareness, and assist government and humanitarian agencies in making informed decisions during emergency response and recovery.
\section*{Future Work}
Although the current system shows strong performance, there are several avenues for improvement and expansion. Ongoing experimentation involves evaluating more advanced architectures such as DeepLabv3+ and transformer-based segmentation models, which are known for their ability to capture long-range spatial dependencies and deliver sharper object boundaries. These models may enhance accuracy in urban regions with dense, overlapping structures.
Another promising direction is the integration of drone-based imagery with satellite data. Drones provide extremely high-resolution, street-level perspectives that complement the broad coverage of satellite images. Combining both sources can enable multi-scale damage analysis from city-wide impact assessment to fine-grained inspection of individual buildings. This hybrid approach would make the system more reliable, especially in regions where satellite images are obstructed by clouds, smoke, or debris.
Future improvements could also include the use of multi-spectral or SAR imagery to increase robustness under adverse environmental conditions. Expanding the training dataset to include more diverse geographical locations and architectural styles would further enhance the model’s generalization ability. On the deployment side, cloud-based processing, GIS map integration, and mobile accessibility could transform the system into a real-time operational tool for emergency teams. Ultimately, the project can grow into a comprehensive disaster analytics platform capable of supporting loss estimation, population exposure analysis, and resource prioritization during large- scale emergencies.

\vspace{12pt}

\end{document}